\documentclass{article}
\usepackage{xcolor}      %
\usepackage{soul}        
\sethlcolor{yellow!30}   
\usepackage{booktabs} 
\usepackage{multirow} 
\usepackage{graphicx} 
\usepackage{tabularx}
\usepackage{makecell}
\usepackage{microtype}
\usepackage{enumitem}
\usepackage{graphicx}
\usepackage{caption}
\usepackage{subcaption}
\usepackage[utf8]{inputenc}
\usepackage{tikz}
\usepackage{pgfplots}
\pgfplotsset{compat=1.17}
\usepackage{xurl}
\usepackage{balance}
\usepackage{flushend}
\usepackage[utf8]{inputenc}
\usepackage{booktabs}
\usepackage{tcolorbox}
\usepackage{enumitem}
\usepackage{xcolor}

\newtcolorbox{promptbox}[1]{
    colback=gray!5,
    colframe=gray!75,
    fonttitle=\bfseries,
    title=#1,
    arc=0mm,
    boxrule=0.5pt,
    left=5pt,
    right=5pt,
    top=5pt,
    bottom=5pt
}
\definecolor{c_prop}{RGB}{70, 130, 180}   
\definecolor{c_open}{RGB}{34, 139, 34}    
\definecolor{c_ours}{RGB}{220, 20, 60}    
\definecolor{c_bg}{RGB}{245, 245, 250}    

\usepackage{hyperref}

\newcommand{\parabf}[1]{\noindent\textbf{#1}}

\usepackage[preprint]{icml2026}

\usepackage{amsmath}
\usepackage{amssymb}
\usepackage{mathtools}
\usepackage{amsthm}
\usepackage{graphicx}
\usepackage{multirow}
\usepackage[table]{xcolor}

\usepackage[capitalize,noabbrev]{cleveref}

\theoremstyle{plain}

\theoremstyle{definition}

\theoremstyle{remark}

\newcommand{\ourmodel}{SVE-ASCII}
\newcommand{\ourbenchmark}{ASCIIArt-Bench}
\newcommand{\ourdataset}{ASCIIArt-7K}
\newcommand{\ascii}{ASCII art}
\usepackage[textsize=tiny]{todonotes}

\icmltitlerunning{Unlocking the Latent Canvas: Eliciting and Benchmarking Symbolic Visual Expression in LLMs}

\begin{document}

\newcommand{\myteaser}{
    \centering
    \includegraphics[width=\linewidth]{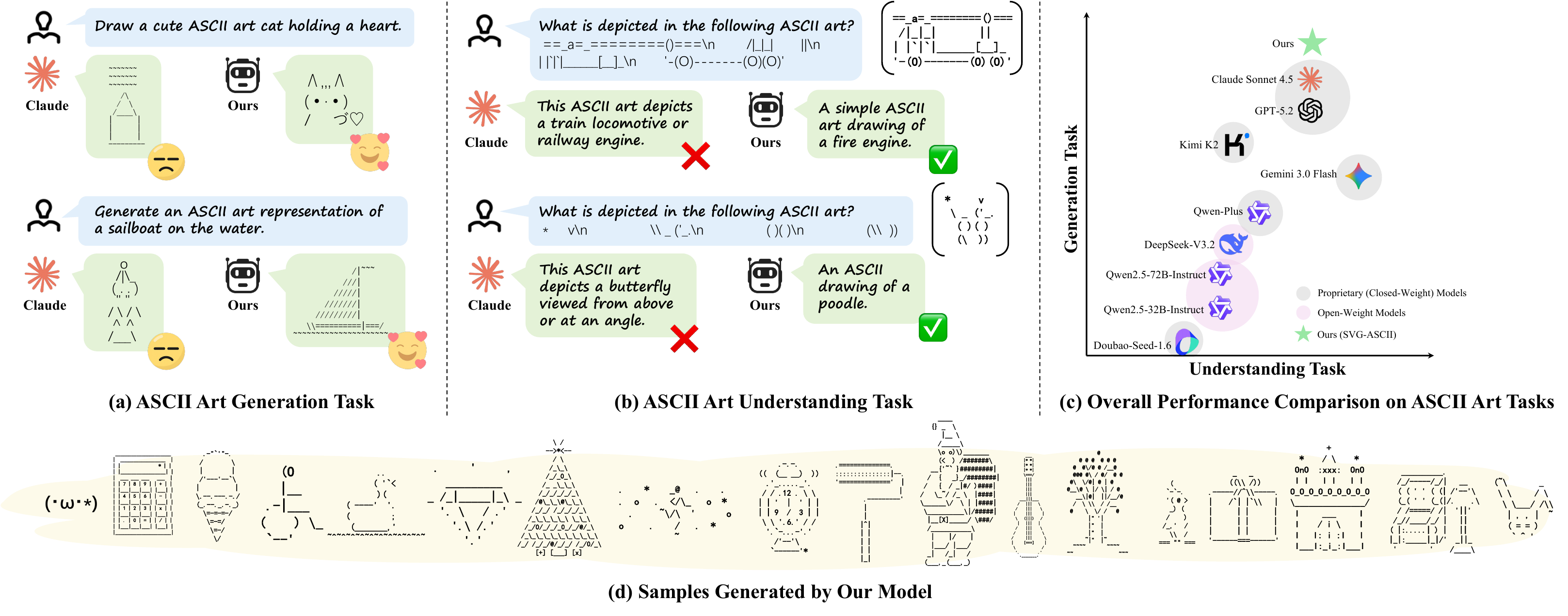}
    \captionof{figure}{\textbf{Capabilities of SVE-ASCII.} Our unified framework achieves high-quality ASCII art generation (a) while accurately interpreting the underlying shapes and semantics (b). Quantitative results in (c) demonstrate that our model significantly outperforms state-of-the-art baselines, and (d) highlights its versatility in synthesizing diverse ASCII art across varying categories and scales.}
    \label{fig:teaser}
}

\twocolumn[
  \icmltitle{Unlocking the Latent Canvas: Eliciting and Benchmarking Symbolic Visual Expression in LLMs}

\vskip 0.12in
\begin{center}
{\large
Yiren Zheng$^{1}$\quad
Shibo Li$^{2}$\quad
Jiaming Liu$^{3}$\quad
Haofan Wang$^{4}$\quad
Yiren Song$^{5\dagger}$
\par}
\vskip 0.06in
{\large
$^{1}$Tsinghua University\quad
$^{2}$Fudan University\quad
$^{3}$Alibaba Group\quad

$^{4}$Lovart AI\quad
$^{5}$National University of Singapore
\par}
\end{center}
\vskip 0.12in
\icmlcorrespondingauthor{Yiren Song}{\phantom{songyiren725@gmail.com}}
  \icmlkeywords{Machine Learning, ICML}
  \vspace{0.3em}
  \begin{center}
  \small
  \textbf{Project Page:}
  \url{https://sve-ascii.github.io}
  \end{center}
  \vskip 0.2in
  \myteaser
  \vskip 0.2in
]



\printAffiliationsAndNotice{$^{\dagger}$Corresponding author}

\begin{abstract}
Current multimodal approaches predominantly treat visual generation as an external process—relying on pixel rendering or code execution—thereby overlooking the native visual representation capabilities latent within Large Language Models (LLMs). In this work, we unlock this potential through ASCII art, a compact, efficient, and text-native visual format. We introduce SVE-ASCII, a unified framework designed to elicit and benchmark Symbolic Visual Expression directly within the pure text space. To address the scarcity of systematic resources, we construct ASCIIArt-7K, a high-quality dataset synthesized via a novel ``Seed-and-Evolve" pipeline that augments human-curated anchors through in-context stylistic editing. We further implement a unified instruction-tuning strategy that jointly optimizes for both Generation (Text-to-ASCII) and Understanding (ASCII-to-Text). Crucially, our experiments reveal a critical phenomenon regarding task duality: while it is established that perception aids generation, we provide compelling evidence that generative training significantly enhances visual comprehension. This confirms a mutually reinforcing cycle in symbolic visual processing—a relationship previously hypothesized but rarely empirically proven in the visual domain. We release our dataset, the ASCIIArt-Bench benchmark, and the SVE-ASCII model, establishing a robust baseline for native text-based visual intelligence.

\end{abstract}

\section{Introduction}
\label{submission}

Large Language Models (LLMs) have demonstrated exceptional reasoning capabilities within the textual domain. However, bridging the gap between \textit{textual semantics} and \textit{visual representation} remains a fundamental challenge. Current solutions typically bypass the LLM's native output space, relying on ``mediated'' representations such as programmatic code (e.g., SVG, Matplotlib) \cite{OmniSVG, Chat2SVG, VisualChatGPT}or discrete visual tokens \cite{DBLP:conf/nips/KohFS23, DBLP:conf/bigdataconf/SeghairBAB24,DBLP:conf/acl/ChakrabartySWPY23}that require heavy detokenization. While effective, these methods decouple visual reasoning from the LLM’s inherent text-generation process, treating visual tasks as auxiliary outputs rather than an integrated cognitive ability.

We argue that LLMs possess a latent, native capacity for visual expression that does not require external rendering engines. ASCII art represents a unique, \textit{isomorphic} middle ground: it encodes 2D spatial structure using standard 1D characters. This format is compact, structurally explicit, and computationally efficient, making it an ideal medium to investigate Symbolic Visual Expression (SVE) directly within the text space. Despite this potential, ASCII art has largely been treated as a novelty or a classification target in previous literature \cite{ASCIIEval2024,ASCIIBench}, lacking a systematic framework for generative modeling and holistic understanding.

To bridge this gap, we introduce \textbf{SVE-ASCII}, a unified framework that reformulates visual processing as a bidirectional language modeling task. A major bottleneck in this domain is the scarcity of high-quality, aligned data. To address this, we propose a scalable synthesis pipeline termed Seed-and-Evolve. This approach starts with high-quality, human-curated seed samples and expands them using an In-Context Learning (ICL) based editing strategy, ensuring both structural integrity and semantic diversity. The resulting dataset, \textbf{ASCIIArt-7K}, serves as a robust foundation for instruction tuning.

Leveraging this data, we train a unified model to perform both ASCII art Generation (Text $\to$ Image) and ASCII Art Understanding (Image $\to$ Text). This joint training regime leads to our most significant empirical finding: \textbf{The Cycle of Mutual Reinforcement}. In the field of unified modeling \cite{Show-o,DBLP:conf/cvpr/LuCL0KMHK24,DBLP:journals/corr/abs-2512-16584}, it is widely accepted that perception (understanding) capabilities lay the foundation for generation \cite{DBLP:conf/nips/TianFICK23,DBLP:conf/iccv/LiPDBP23}. However, the inverse—\textit{whether generation actively benefits understanding}—remains an open question in visual domains. Our results confirm this bidirectional synergy: the rigor required to generate coherent ASCII structures forces the model to attend to fine-grained spatial details, which in turn drastically improves its ability to interpret and classify visual inputs.

Our contributions are threefold:
\begin{itemize}[leftmargin=1.2em, itemsep=2pt, topsep=2pt, parsep=0pt]
    \item \textbf{Task \& Perspective:} We propose \textit{Symbolic Visual Expression} (SVE) as a native visual task for LLMs, utilizing ASCII art to bypass the syntax-semantics gap inherent in code-based or pixel-based generation.
    \item \textbf{Methodology:} We introduce a robust data synthesis pipeline (\textit{Seed-and-Evolve}) and a unified training framework that establishes a new state-of-the-art on our proposed \textbf{ASCIIArt-Bench}.
    \item \textbf{Key Insight:} We empirically demonstrate that generation and understanding are not merely compatible but mutually reinforcing. Specifically, we provide the first systematic evidence that generative training significantly enhances visual comprehension in symbolic representations.
\end{itemize}

\section{Related Works}

\parabf{Visual Synthesis Beyond Pixels.} High-fidelity generation via pixel-based diffusion \cite{SDXL, Imagen} or visual tokens \cite{DBLP:conf/iclr/Chen0CPPSGGMB0P23, DBLP:conf/cvpr/EsserRO21} relies on heavy detokenization decoupled from LLMs. Alternatively, generating visual programs (SVG/Python) \cite{StarVector, IconShop, DeepSVG, Chat2SVG, ViperGPT,DBLP:journals/corr/abs-2509-21404, song2022cliptexture, song2022clipfont, song2023clipvg, song2025layertracer} introduces a ``syntax-semantics gap", where minor errors break rendering. In contrast, we adopt \ascii{} as a native text-space representation. Sharing the same generation mechanism as language, it enables ``what-you-type-is-what-you-see" synthesis without external compilation, eliminating the misalignment between symbolic syntax and visual semantics.

\parabf{ASCII Art Datasets and Benchmarks.} Existing symbolic visual benchmarks are fragmented. ASCIIEval \cite{ASCIIEval2024} establishes rigorous perception protocols but ignores generation, while generative studies \cite{DBLP:journals/corr/abs-2503-14375} often lack semantic diversity or abstraction. To bridge this, we introduce \ourdataset{}, a large-scale dataset preserving stylistic consistency, and \ourbenchmark{}, a unified suite for generation and understanding. Unlike prior disjointed efforts, our benchmark supports both fine-grained structural synthesis and precise semantic interpretation, providing a holistic testbed for symbolic visual modeling in pure text space.

\parabf{Unified Generation and Understanding.} Unified models, which integrate visual understanding and generation within a single framework, have recently emerged as an important research direction \cite{Show-o, de2006bagel, zhang2026sigma, ye2025loom}. In Code Intelligence, generation and understanding are mutually reinforcing, as seen in CodeT5 \cite{DBLP:conf/emnlp/0034WJH21} and CodeBERT \cite{DBLP:conf/emnlp/FengGTDFGS0LJZ20}. However, the visual domain typically fractures these into distinct architectures (e.g., Diffusion vs. ViTs). We unify them at the supervision level by exploiting the isomorphic nature of ASCII art and language. Our experiments demonstrate a mutual reinforcement: improved understanding enhances structural stability, while generative training strengthens symbolic interpretation, effectively extending ``code duality" to the visual domain.

\section{Method}
\label{sec:method}

\begin{figure*}[htbp]
    \centering
    \includegraphics[width=\textwidth]{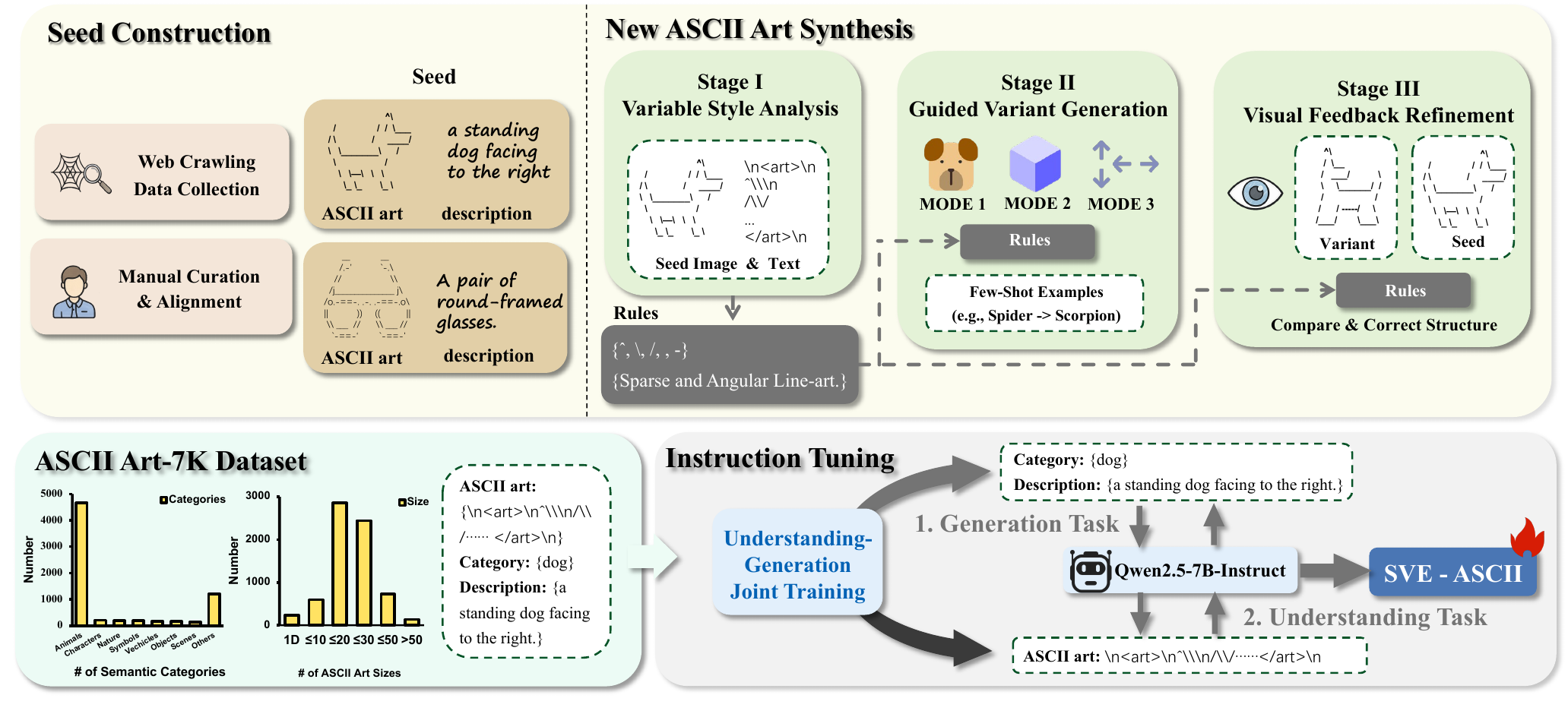}
    \caption{\textbf{Overview of \ourmodel.} First, we construct the \ourdataset{} using a scalable synthesis pipeline. Subsequently, we fine-tune Qwen2.5-7B-Instruct via Understanding-Generation Joint Training. By switching supervision signals between tasks, our model achieves efficient bidirectional capabilities in both understanding and generating ASCII art.}
    \label{fig:overview}
\end{figure*}


We construct \textbf{ASCII Art-7K }, a high-quality dataset for text-based visual art generation in \emph{text space}. Our goal is to obtain diverse and aesthetically pleasing ASCII art samples paired with (i) a natural-language description and (ii) the corresponding ASCII art.

\subsection{Dataset Construction}

\paragraph{Seed Construction.}
We collect raw ASCII art candidates from publicly accessible internet sources, including text-based visual art webpages and user comments. User identifiers and metadata are removed to prevent linkage to specific individuals.

We refine the raw crawl by removing artifacts such as near-duplicates and corrupted layouts, retaining only samples that meet strict criteria for structural integrity and safety. Following this filtration, each ASCII sample is paired with a descriptive label to achieve robust data alignment. These curated pairs function as the seed data for our automated expansion pipeline, which leverages the multimodal capabilities of Gemini~3~Pro to scale the dataset.

\begin{figure}[h]
    \centering
    \includegraphics[width=\linewidth]{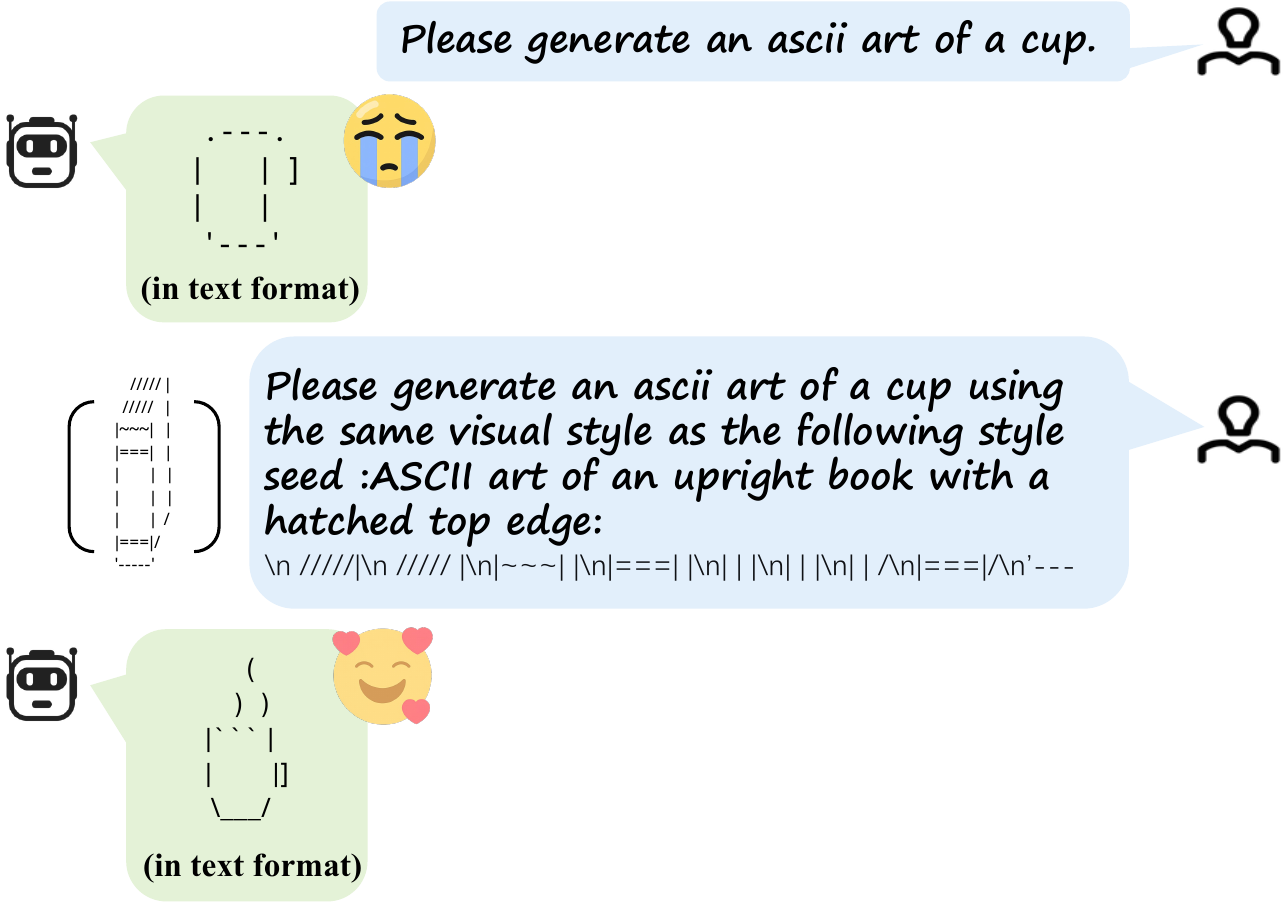}
    \caption{\textbf{Generation from Scratch vs. Imitation.}The quality of the generated ASCII art cup improves significantly when an imitation example is provided in the prompt.}
    \label{fig:model_compare} 
\end{figure}

In preliminary experiments, as shown in \cref{fig:model_compare}, we observe that large language models exhibit limited capability in synthesizing high-quality ASCII-art from scratch, often producing visually incoherent or structurally erroneous patterns. 
However, the model exhibits better performance when imitating existing ASCII-art samples.

Accordingly, we expand the dataset by generating new samples through evolving seed instances while preserving their stylistic structure. 
To operationalize this process, we propose a three-stage pipeline.

\paragraph{Stage I: Variable Style Analysis.}
To capture stylistic essence, we explicitly convert the seed ASCII-art $a_i$ into a rendered image tensor $I_i = \mathcal{R}(a_i)$. This visual representation, alongside the raw text $t_i$, is processed by a multimodal encoder to extract style rules: 
\begin{equation}
    s_i = \mathcal{E}_{\text{mm}}(I_i, t_i).
\end{equation}
The output $s_i$ defines explicit constraints comprising: (i) a character palette $C_p$ (vocabulary constraint), and (ii) structural logic $S_l$ (geometric topology constraint), as exemplified in \cref{tab:style_rules}. These rules are injected into the system prompt to enforce stylistic consistency, transforming generation into a rule-guided synthesis.

\begin{table}[htbp]
    \centering
    \small
    \caption{\textbf{Style Rules Decomposition.} Taking a dog ASCII art as an example, this stage extracts the specific set of characters used and the topological features describing the dog's shape.}
    \label{tab:style_rules}
    
    \renewcommand{\tabularxcolumn}[1]{m{#1}}
    
    \begin{tabularx}{\columnwidth}{@{} >{\centering\arraybackslash}m{1.5cm} c X @{}} 
        \toprule
        \textbf{Component} & \textbf{Input ($I_i, t_i$)} & \textbf{Output(rules)}\\
        \midrule
        
        \makecell{\textbf{Character}\\\textbf{Palette}($C_p$)} 
        &
        \multirow{4}{*}{\begin{minipage}{2.8cm}
            \centering
            \includegraphics[width=1.3cm]{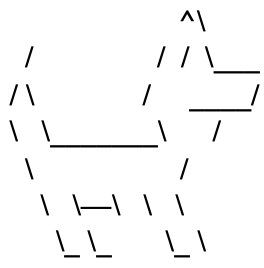}%
            \hfill
            \begin{minipage}[b]{1.2cm} 
                \tiny\ttfamily
               \textbackslash n<art>\textbackslash n\\
                \^{}\textbackslash\textbackslash\textbackslash n\\
                /\textbackslash\textbackslash / \_\_\_/\textbackslash n\\\textbackslash\textbackslash\_\textbackslash\textbackslash\_\textbackslash n\\
                \dots
                
                </art>\textbackslash n
            \end{minipage}
        \end{minipage}}
        &
        \textbf{Boundary/Textures Glyphs:} \par
        \{\texttt{\^{}}, \texttt{\textbackslash}, \texttt{/}, \texttt{\_}, \texttt{-}\}
        \\ 
        \\
        

        
        \makecell{\textbf{Structural}\\\textbf{Logic}($S_l$)} 
        & 
        & 
        \textbf{Topology:} \par
        \{Sparse and Angular Line-art.\}
        \\
        \bottomrule
    \end{tabularx}
\end{table}

\paragraph{Stage II: Guided Variant Generation.}

\begin{figure}[htbp]
    \centering
    \includegraphics[width=1\linewidth]{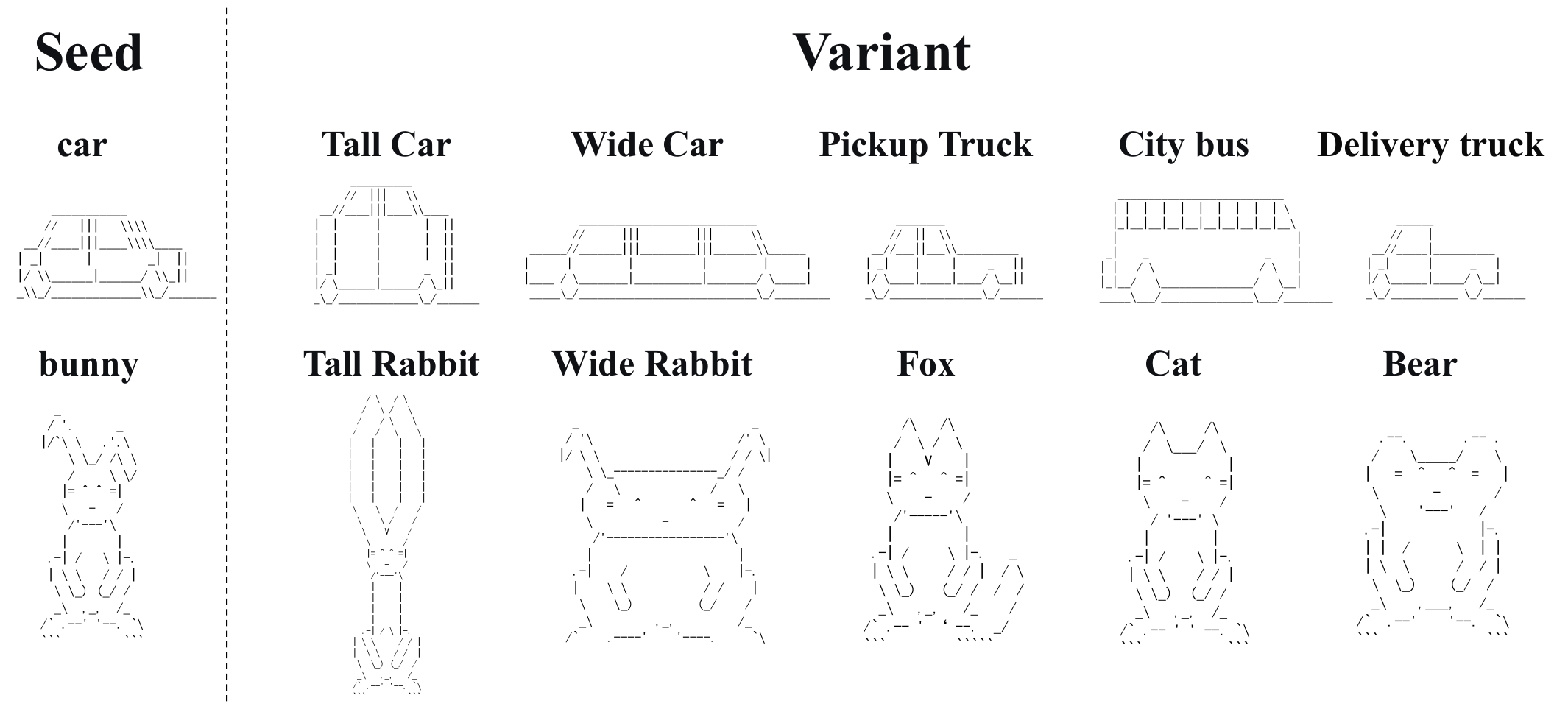}
    \caption{\textbf{Impact of Topology on Style Transfer.} 
     We synthesize variants for different seeds. Visually insensitive subjects (e.g., trucks) demonstrate robust style transfer across diverse shapes. Conversely, visually sensitive subjects (e.g., rabbits) struggle with large morphological changes, often failing to transform into distinct animals such as foxes or cats.}
    \label{fig:variant mode}
\end{figure}

We synthesize variants $a_i^{(k)}$ by constructing a system prompt conditioned on the seed tuple and extracted style rules $s_i$. As illustrated in \cref{fig:variant mode}, style transfer quality is contingent on topological rigidity. We identify visually sensitive subjects (e.g., organic animals) that resist shape migration, because the artistic style is highly dependent on structural outlines—where the stylistic rendering is tailored specifically to the subject's morphology, 
versus visually insensitive objects (e.g., rigid vehicles) that decouple style from geometry.
Accordingly, we apply adaptive structured evolution modes $m_k$:
\begin{itemize}[leftmargin=*, nosep]
    \item Micro-perturbation ($m_{\text{micro}}$) for sensitive entities, restricting changes to local attributes;
    \item Creative variant ($m_{\text{creative}}$) for insensitive objects, allowing category-level transformation;
    \item Global structural variant ($m_{\text{global}}$) such as tall or wide stretching, is universally applicable.
\end{itemize}
To further enforce style locking, we retrieve multimodal few-shot examples $\mathcal{F}$ as visual demonstrations. The process is modeled as:\begin{equation}a_i^{(k)} \sim p(a \mid s_i, a_i, \mathcal{F}, m_k).\end{equation}

\paragraph{Stage III: Visual Feedback Refinement.}
Guided generation may yield minor structural deviations (e.g., misalignment). To mitigate this, we introduce a refinement module that corrects the draft $\hat{a}_i^{(k)}$ under explicit style constraints.
Specifically, we verify the rendered draft $\hat{I}_i^{(k)} = \mathcal{R}(\hat{a}_i^{(k)})$ against the seed image $I_i$. The final output $a_i^{(k)}$ is obtained by localizing and correcting visual discrepancies:
\begin{equation}
a_i^{(k)} = \mathcal{G}_{\text{review}}(I_i, \hat{I}_i^{(k)}, s_i, m_k),
\end{equation}
where $\mathcal{G}_{\text{review}}(\cdot)$ denotes a multimodal refinement operator. 
Crucially, this refinement is strictly conditioned on the style rules $s_i$ to prevent stylistic drift during structural correction. 
The refined ASCII art $a_i^{(k)}$ is taken as the final synthesized output of Stage~III and paired with its corresponding instruction to form an entry in the augmented dataset. 
Each entry consists of a text-format ASCII art, its associated category, and a natural language description.

\subsection{\ourbenchmark}
\label{subsec:benchmark}

To address the lack of standardized evaluations for symbolic visual processing in text space, we introduce \textbf{ASCIIArt-Bench}. This benchmark comprises 400 samples evenly divided with 200 dedicated to generation and 200 to understanding. By rigorously minimizing overlap with training data, the benchmark is designed to ensure the assessment of genuine generalization and faithfully validate the model's practical usability and robustness in real-world scenarios.

\parabf{Generation Task.}
This task evaluates the model's ability to synthesize a diverse range of symbolic visual expressions within a pure text space. Given instruction or description, the model is required to generate the corresponding text-format ASCII art. 
This task focuses on whether the model can correctly map textual semantics and structural constraints into coherent symbolic layouts. 
To distinguish between memorization and generalization, we construct two evaluation subsets: 
(i) a Recall Subset, containing instructions sampled directly from the training set; 
(ii) a Generalization Subset, consisting of novel instructions synthesized by Gemini-3-Pro that are semantically distinct from the training prompts.

\parabf{Understanding Task.}
The understanding task evaluates a model’s ability to interpret symbolic visual expressions represented in text space. Given an ASCII art input, the model is required to infer its semantic meaning under two complementary settings: 
(i) Direct Recognition (generating the label through open-ended inference), and (ii) Category Selection (identifying the correct label from a predefined candidate set).


To disentangle memorization from generalization, we include samples from unseen categories and unseen instances of seen categories, enabling separate assessment of category-level and instance-level generalization.

All detailed task prompt templates and benchmark configurations are provided in Appendix~\ref{benchmark-task}.

\subsection{Fine Tuning}

We perform instruction tuning on \ourdataset, where each ASCII art instance is paired with both generation and understanding supervision.
Specifically, the same ASCII sample can be used either as a target for generation (text $\rightarrow$ ASCII) or as an input for understanding (ASCII $\rightarrow$ text/category), by switching the supervision signal.
This unified formulation enables joint learning of symbolic visual creation and interpretation without introducing additional data.

\parabf{Problem Definition.}
We consider a unified instruction-following formulation over symbolic visual data.
Each training sample is represented as a triplet $(x, y, m)$, where $m$ specifies the task mode: generation or understanding.

For generation, given a natural-language instruction or description $x$ (e.g., ``a cute crying face'' or ``a cat wearing sunglasses''), the model produces an text-format ASCII art artifact $y$:
\begin{equation}
    y = (t_1, t_2, \ldots, t_T), \quad t_j \in \mathcal{V},
\end{equation}
where $\mathcal{V}$ is the ASCII art vocabulary and newline tokens define row breaks. For understanding, the input $x$ contains an ASCII art instance, and the target $y$ is either a natural-language description or a semantic category label. We model both tasks using a shared autoregressive policy $\pi_\theta(y \mid x)$ and seek outputs that are (i) aligned with the input semantics, (ii) structurally coherent, and (iii) aesthetically pleasing.

\parabf{Instruction-Tuning.}
We train the model with maximum-likelihood learning over the merged dataset $\mathcal{D}$:
\begin{equation}
\mathcal{L}_{\text{SFT}}(\theta)
= - \mathbb{E}_{(x,y)\sim\mathcal{D}}
\left[\sum_{t=1}^{|y|} \log \pi_\theta(y_t \mid x, y_{<t})\right].
\end{equation}
This objective teaches the model both to synthesize structured ASCII art from textual descriptions and to interpret symbolic visual patterns into semantic representations.


\section{Evaluation}
\label{sec:evaluation}

\begin{table*}[t!]
\centering
\caption{\textbf{Main results on our ASCII Art Benchmark comparing Proprietary (Closed-Weight) and Open-Weight models.} Consistent with the benchmark definition, we evaluate models on the Recall Subset (In-Distribution) and the Generalization Subset (Out-of-Distribution). The metrics reported are Semantic Alignment (SA), Instruction Faithfulness (IF), Structural Coherence (SC), Spatial Logic (SL), Character Efficiency (CE), and the Weighted Composite Score (Comp.). \textbf{Bold} indicates the best performance, and \textcolor{teal}{blue text} indicates the second best. \textbf{Inst.} denotes the instruction-tuned version of the model, and Qwen2.5-7B-Inst. refers to the base model in this benchmark.}
\label{tab:main_results}

\resizebox{\textwidth}{!}{%
\begin{tabular}{l|cccccc|cccccc}
\toprule
\multirow{2}{*}{\textbf{Model}} & \multicolumn{6}{c|}{\textbf{Recall Subset (In-Distribution)}} & \multicolumn{6}{c}{\textbf{Generalization Subset (Out-of-Distribution)}} \\ 
\cmidrule(lr){2-7} \cmidrule(lr){8-13}
 & SA & IF & SC & SL & CE & \textbf{Comp.} & SA & \textbf{IF}$^\dagger$ & SC & SL & CE & \textbf{Comp.} \\ 
\midrule
GPT-5.2 & 0.558 & 0.544 & 0.785 & 0.690 & 0.825 & 0.633 & 0.604 & 0.568 & 0.756 & 0.713 & 0.801 & 0.650 \\
Kimi K2 & 0.383 & 0.420 & 0.812 & 0.607 & 0.814 & 0.537 & 0.449 & 0.392 & 0.808 & 0.596 & 0.789 & 0.539 \\
Doubao-Seed-1.6 & 0.259 & 0.265 & 0.694 & 0.450 & 0.838 & 0.413 & 0.332 & 0.347 & 0.698 & 0.476 & \textcolor{teal}{0.845} & 0.465 \\
Qwen-Plus & 0.346 & 0.351 & 0.804 & 0.577 & 0.836 & 0.500 & 0.398 & 0.371 & 0.782 & 0.595 & 0.834 & 0.519 \\
Gemini 3.0 Flash & 0.474 & 0.453 & 0.544 & 0.585 & 0.754 & 0.522 & 0.490 & 0.403 & 0.535 & 0.590 & 0.723 & 0.505 \\
Claude Sonnet 4.5 & \textcolor{teal}{0.692} & \textcolor{teal}{0.701} & \textcolor{teal}{0.825} & \textcolor{teal}{0.754} & \textcolor{teal}{0.871} & \textcolor{teal}{0.742} & \textcolor{teal}{0.710} & \textbf{0.630} & \textcolor{teal}{0.817} & \textbf{0.770} & \textbf{0.864} & \textbf{0.722} \\
\midrule
DeepSeek-V3.2 & 0.390 & 0.345 & 0.707 & 0.519 & 0.718 & 0.474 & 0.458 & 0.351 & \textbf{0.821} & 0.591 & 0.789 & 0.528 \\
Qwen2.5-72B-Inst. & 0.343 & 0.323 & 0.808 & 0.533 & 0.853 & 0.485 & 0.325 & 0.288 & 0.770 & 0.476 & 0.825 & 0.451 \\
Qwen2.5-32B-Inst. & 0.330 & 0.309 & 0.779 & 0.546 & 0.858 & 0.475 & 0.314 & 0.280 & 0.723 & 0.495 & 0.818 & 0.441 \\
Qwen2.5-7B-Inst. & 0.238 & 0.251 & 0.723 & 0.412 & 0.805 & 0.398 & 0.225 & 0.211 & 0.609 & 0.354 & 0.679 & 0.342 \\

\midrule
\textbf{\ourmodel{}} & \textbf{0.936} & \textbf{0.946} & \textbf{0.892} & \textbf{0.926} & \textbf{0.918} & \textbf{0.929} & \textbf{0.719} & \textcolor{teal}{0.619} & 0.762 & \textcolor{teal}{0.733} & 0.824 & \textcolor{teal}{0.703} \\
\bottomrule
\end{tabular}%
}
\begin{flushleft}
\footnotesize{$^\dagger$ \textbf{IF (Instruction Faithfulness)} is the critical metric for the Generalization Subset, measuring robustness to novel prompts.}
\end{flushleft}
\end{table*}
\begin{table}[t]
\centering
\setlength{\tabcolsep}{3pt} 
\renewcommand{\arraystretch}{1.1} 

\caption{\textbf{Results on the ASCII Art Understanding Benchmark comparing Proprietary (Closed-Weight) and Open-Weight models.} We evaluate models on two tasks: Task I: Direct Understanding and Task II: Category Selection. All metrics are reported as Accuracy (\%), evaluating performance on Seen, Unseen, and Overall. Bold indicates the best performance, and \textcolor{teal}{blue text} indicates the second best.}
\label{tab:understanding_results}

\resizebox{\columnwidth}{!}{%
\begin{tabular}{l|ccc|ccc}
\toprule
\multirow{2}{*}{\textbf{Model}} & \multicolumn{3}{c|}{\textbf{Task I (Direct)}} & \multicolumn{3}{c}{\textbf{Task II (Select)}} \\
\cmidrule(lr){2-4} \cmidrule(lr){5-7}
 & Seen & Unseen & \textbf{Avg.} & Seen & Unseen & \textbf{Avg.} \\
\midrule
GPT-5.2 & 11.0 & \textcolor{teal}{17.0} & 14.0 & 18.0 & 17.0 & 17.5 \\
Kimi-K2 & 9.0 & 6.0 & 7.5 & 10.0 & 5.0 & 7.5 \\
Doubao-Seed-1.6 & 2.0 & 4.0 & 3.0 & 5.0 & 4.0 & 4.5 \\
Qwen-Plus & 9.0 & 10.0 & 9.5 & 8.0 & 6.0 & 7.0 \\
Claude-Sonnet-4.5 & 16.0 & 14.0 & 15.0 & \textcolor{teal}{19.0} & \textcolor{teal}{21.0} & \textcolor{teal}{20.0} \\
Gemini-3.0-Flash & \textbf{37.0} & \textbf{24.0} & \textbf{30.5} & \textbf{42.0} & \textbf{35.0} & \textbf{38.5} \\
\midrule
DeepSeek-V3.2 & 6.0 & 10.0 & 8.0 & 10.0 & 4.0 & 7.0 \\
Qwen2.5-72B-Inst. & 6.0 & 6.0 & 6.0 & 7.0 & 7.0 & 7.5 \\
Qwen2.5-32B-Inst. & 5.0 & 5.0 & 5.0 & 10.0 & 4.0 & 7.0 \\
Qwen2.5-7B-Inst. & 3.0 & 4.0 & 3.5 & 2.0 & 3.0 & 2.5 \\
\midrule
\textbf{\ourmodel{}} & \textcolor{teal}{29.0} & 11.0 & \textcolor{teal}{20.0} & 16.0 & 5.0 & 10.5 \\
\bottomrule
\end{tabular}%
}
\end{table}

\subsection{Experimental Setups}
\label{subsec:training_settings}

\parabf{Instruction-Tuning Implementation.}
We adopt Qwen2.5-7B-Instruct \cite{Qwen2.5} as the base model. We perform instruction-tuning to adapt the base model to the \ascii{} generation and understanding task.
All experiments are conducted using full-parameter fine-tuning.
We employ the AdamW optimizer with an initial learning rate of 1e-5.
Training is conducted for 10 epochs with a global batch size of 2, distributed across 8 GPUs.
All instruction tuning experiments are implemented using the LLaMA-Factory framework and conducted on 8 $\times$ NVIDIA A800 GPUs.
Under this configuration, the entire training process requires approximately 2 hours.

\parabf{Metrics.} For the \ascii{} generation task in \ourbenchmark{} (\ref{subsec:benchmark}), we adopt a reference-free, multi-modal evaluation protocol using a strong Vision-Language Model (Gemini 3.0 Pro) as an automated judge. The judge takes both the \textit{rendered image} and the \textit{raw text} as input to evaluate the generation quality across five distinct dimensions:
\begin{itemize}[leftmargin=*, nosep]
    \item Semantic Alignment (SA): Evaluates whether the generated art visually represents the core object described in the prompt (e.g., is it recognizable as a dog?).
    \item Instruction Faithfulness (IF): Evaluates the model's adherence to specific constraints, attributes, or perturbations (e.g., ``facing left", ``add a star"). This is the most critical metric for assessing robustness on our Generalization Subset.
    \item Structural Coherence (SC): Evaluates the visual integrity of the art by detecting disconnected lines or broken contours.
    \item Spatial Logic (SL): Evaluates the anatomical or geometrical correctness of component placement (e.g., legs below the body).
    \item Character Efficiency (CE): Evaluates the cleanliness of the generation by identifying visual noise, repetitive character spam, or unnecessary artifacts.
\end{itemize}

To provide a holistic ranking, we compute a weighted Composite Score (Comp.) by aggregating the five evaluation dimensions with explicitly assigned weights.We assign the highest weight to IF (0.35), as the core objective of our benchmark is to evaluate whether a model can correctly follow fine-grained instructions and structural constraints. SA is assigned a weight of 0.25. While semantic correctness is essential, it is less challenging than precise constraint execution and therefore receives a slightly lower weight than IF. SC and SL are each assigned a weight of 0.15. These two criteria measure complementary aspects of structural quality. Finally, CE is assigned a weight of 0.10. They are secondary to semantic correctness and structural validity in our setting.This weighting scheme reflects our emphasis on instruction adherence and semantic fidelity, while still accounting for structural soundness and stylistic cleanliness in evaluating overall ASCII art quality.

For the two aspects of the \ascii{} understanding task in \ourbenchmark{} (\ref{subsec:benchmark}), we also employ an LLM-based (Gemini 3.0 Pro) judge to robustly parse and evaluate the model outputs.
Specifically, the judge determines whether the generated description is semantically consistent with the ground truth in the direct understanding task, and whether the predicted category matches the ground-truth label in the category selection task.Both evaluations are formulated as binary decisions, and we report the corresponding accuracy as the final metric.

The LLM's judge prompt template of generation task and understanding task can be find in Appendix~\ref{LLM-as-Judge}

\parabf{The experiments of LLM as judge.} 
To design and calibrate the LLM-based judge for the two tasks in \ourbenchmark{}, we aggregated a pool of 2,000 evaluation samples by collecting outputs from five representative models (GPT-5.2, Qwen2.5-7B-Instruct, Gemini 3.0 Flash, Claude Sonnet 4.5, and Kimi K2) across the benchmark's 400 tasks (comprising 200 generation and 200 understanding tasks). From this pool, we randomly selected a diverse subset of 200 samples for manual annotation.

To establish high-quality ground truth, each sample in this subset was independently scored by three human annotators, with the final reference scores derived by averaging the three ratings to mitigate individual subjectivity.
We then benchmarked multiple candidate models (including Gemini 3 Pro, Claude Sonnet 4.5, and GPT-5.2) using iteratively refined prompts to maximize alignment with these human scores. On the held-out validation set, Gemini 3 Pro demonstrated superior performance, achieving a Pearson correlation of 0.85 and a Spearman correlation of 0.87 (MSE = 0.064). In comparison, other strong candidates like Claude Sonnet 4.5 (Pearson: 0.80, Spearman: 0.74, MSE: 0.091) and GPT-5.2 (Pearson: 0.75, Spearman: 0.66, MSE: 0.086) exhibited weaker alignment with human judgments.
This strong consistency confirms that Gemini 3 Pro serves as a reliable proxy for human evaluation; consequently, we adopted this validated judge for all subsequent experiments.


\begin{figure*}[htbp]
    \centering
    \includegraphics[width=\textwidth]{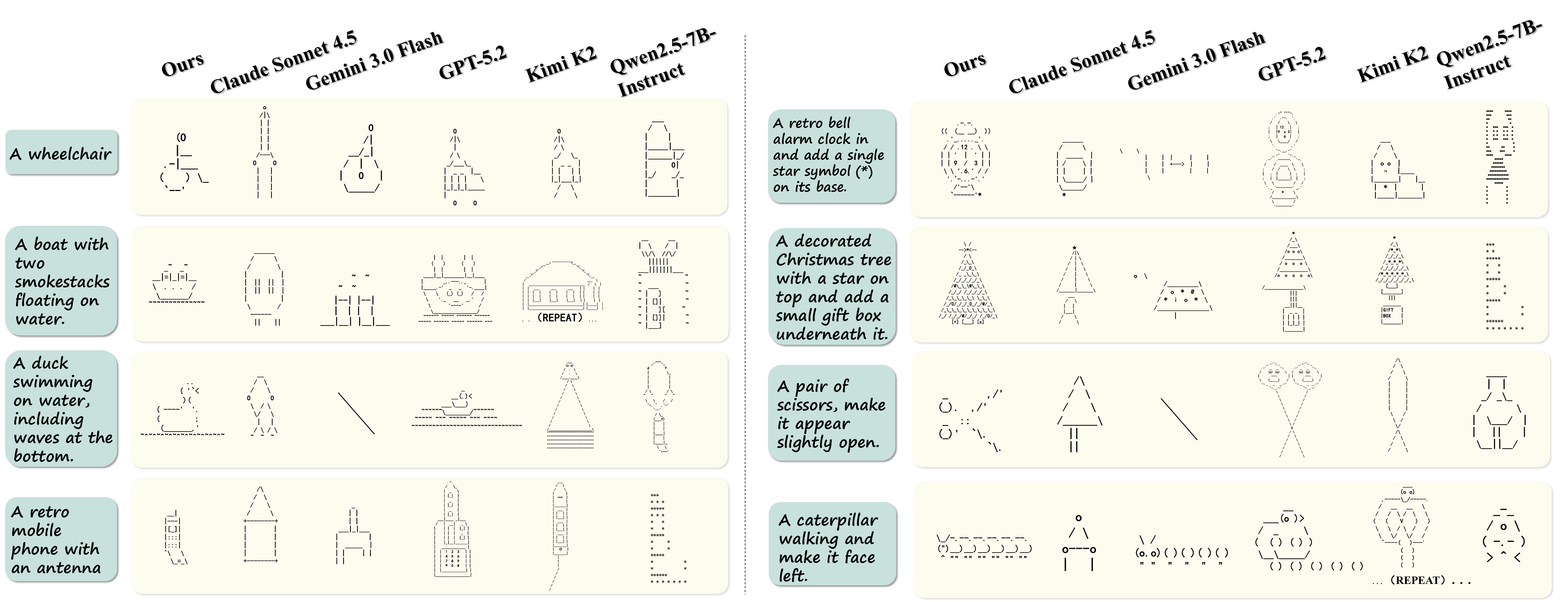}
    \caption{\textbf{Qualitative comparison with SOTA methods on the Generation Task.} We compare the proposed method with SOTA baselines on our evaluation benchmarks, presenting results from the Recall Subset on the left and the Generalization Subset on the right.}    
    \label{fig:Qualitative Evaluations}
\end{figure*}

\parabf{Baselines.}
To evaluate the effectiveness of our proposed method, we benchmark against a comprehensive set of strong baselines categorized into two groups: Proprietary Models and Open-Weights Models.

\noindent\textit{Proprietary (Closed-Weight) Models.}
We select widely adopted commercial LLMs known for their superior performance in reasoning, coding, and instruction following. These include GPT-5.2, Claude Sonnet 4.5~\cite{Claude}, Gemini 3.0 Flash~\cite{DBLP:journals/corr/abs-2312-11805}, Qwen-Plus~\cite{Qwen2.5}, Grok-4, Kimi K2~\cite{DBLP:journals/corr/abs-2507-20534}, and Doubao-Seed-1.6.
These models serve as the ``upper bound" of general-purpose capabilities without task-specific fine-tuning. Specifically, Claude Sonnet 3.5 is included as a critical competitor due to its recognized strengths in spatial reasoning and complex code generation, which are closely related to ASCII art synthesis.

\noindent\textit{Open-Weight Models.}
To investigate the scaling behavior of ASCII art generation and isolate the contribution of our instruction tuning, 
We evaluate the DeepSeek-V3.2~\cite{DBLP:journals/corr/abs-2512-02556} and the Qwen2.5-Instruct series across three model sizes: 7B, 32B, and 72B~\cite{Qwen2.5}.
Comparing against the Qwen2.5-7B-Instruct (Base) allows us to directly quantify the performance gain attributed to our proposed method. Furthermore, including the larger 32B and 72B variants enables us to analyze whether the capability to generate symbolic visual structures emerges naturally with increased parameter scale or strictly requires specialized alignment.

\subsection{Quantitative Evaluations}

We report the evaluation results of \ourmodel{} and baselines on \ourbenchmark{} in Table~\ref{tab:main_results} and Table~\ref{tab:understanding_results}, covering both the generation and understanding tasks.

As shown in Table~\ref{tab:main_results}, \ourmodel{} consistently outperforms most proprietary and open-weight baselines on the generation task across both the Recall Subset and the Generalization Subset. In particular, our model achieves the highest IF on the Recall Subset (In-Distribution), indicating strong capability in following fine-grained constraints rather than merely producing plausible shapes. On the more challenging Generalization Subset (Out-of-Distribution), while Claude Sonnet 4.5 attains the best IF, \ourmodel{} achieves a very close second-best IF and a competitive Composite Score (0.703 vs. 0.722), substantially outperforming other commercial models such as Gemini 3.0 Flash and GPT-5.2. These results demonstrate the robustness of our method under variant instructions with visual perturbations.

For the understanding task (Table~\ref{tab:understanding_results}), \ourmodel{} does not consistently outperform the strongest proprietary models such as Gemini 3.0 Flash or Claude Sonnet 4.5 across all metrics. Instead, it achieves moderate improvements over several competitive baselines and approaches the performance of the top commercial systems on specific settings. In particular, our model outperforms GPT-5.2 and Claude Sonnet 4.5 on the Direct Understanding task, while remaining competitive on Category Selection.
Nevertheless, the overall accuracy remains relatively low for all evaluated models, including ours, indicating that symbolic visual understanding from ASCII art is intrinsically difficult. These results suggest that, although joint generation–understanding training yields measurable gains, robust semantic interpretation of symbolic visual patterns remains an open challenge.

\subsection{Qualitative Evaluations}


We compare our method with representative proprietary baselines using eight distinct text prompts for the ASCII art generation task, as shown in \cref{fig:Qualitative Evaluations}. 
Kimi K2 exhibits severe generation degeneration; while it occasionally captures basic shapes, it frequently collapses into infinite repetitive loops, producing excessively long and meaningless vertical patterns (e.g., in Boat and Caterpillar). Gemini 3.0 Flash suffers from instability, often outputting conversational artifacts (e.g., ``Actually...") or truncated code blocks, indicating a lack of global planning for the canvas. Qwen2.5-7B-Instruct and Claude Sonnet 4.5 struggle with semantic alignment, producing either cluttered character blocks or overly abstract geometric shapes that fail to represent the target objects (e.g., Wheelchair and Scissors). 
GPT-5.2, while capable of retrieving details, often results in visual clutter due to misalignment and over-complexity. 

In contrast, \ourmodel{} consistently outperforms all baselines. It not only generates high-fidelity ASCII art with accurate geometric structures but also excels in visual completeness, precise character selection, and spatial layout. Consequently, the symbolic graphics generated by our method convey significantly greater semantic richness compared to other approaches. Our \ourmodel{} also demonstrates superior instruction-following capabilities, successfully handling fine-grained constraints (e.g., adding a star to the Alarm Clock) and directional control. This consistent high-quality performance across both seen and unseen scenarios effectively validates the model's robustness and practical usability in real-world applications.

\subsection{Effect of Understanding-Generation Joint Training and Discussion}
\label{subsec:ablation}
To evaluate the effect of joint training,
we compare \ourmodel{} with models trained in a single-task setting, i.e., using only the generation task or only the understanding task. All models share the same backbone architecture (Qwen2.5-7B-Instruct), are trained with identical optimization settings, including the number of training steps, batch size, learning rate.
The results in Table~\ref{tab:table3} indicate a clear mutual benefit between the two tasks.

\parabf{Q1: Does understanding improve generation performance?}
As shown in \cref{tab:table3}, Part I, adding the understanding task significantly improves generation quality. Specifically, IF increases by 12.5\% on the Recall Subset and 12.8\% on the Generalization Subset.
We believe this is because the understanding task teaches the model to ``read" and analyze spatial structures. By learning to identify what makes an ASCII art correct (e.g., recognizing categories or attributes), the model becomes a better ``judge" of its own output, allowing it to follow complex instructions more primarily.

\parabf{Q2: Does generation enhance understanding ability?}
Conversely, the generation task greatly boosts the model's ability to interpret ASCII art (Part II). The unified model achieves a massive 42.9\% relative improvement in the Direct Understanding task.
We attribute this to the fact that generating ASCII art requires the model to construct images character by character. This process forces the model to pay attention to every detail of the visual patterns. This ``constructive" experience makes it much easier for the model to recognize and describe ASCII art when it sees it.

\parabf{Summary of Findings.}
These results confirm that generation and understanding are not separate skills. Learning to generate helps the model understand details, while learning to understand helps the model follow structural rules. The \ourmodel{} successfully combines these strengths to achieve the best overall performance.
\begin{table}[t]
\centering
\caption{\textbf{Effect of Understanding-Generation Joint Training.} We compare \ourmodel{} with models trained in a single-task setting. Set A: Recall Subset. Set B: Generalization Subset. $^\dagger$ \textbf{IF (Instruction Faithfulness)} is the critical metric for the Generalization Subset, measuring robustness to novel prompts. $\Delta$ denotes the relative improvement percentage.}
\label{tab:ablation_multitask}
\resizebox{\columnwidth}{!}{%
\begin{tabular}{lcccc}
\toprule
\textbf{Metric / Task} & \textbf{Single-Task} & \textbf{\ourmodel{}} & \textbf{$\Delta$ (\%)} \\
\midrule
\multicolumn{4}{l}{\textit{\textbf{Part I: Generation Task (Composite Score)}}} \\
\midrule
Set A: Semantic Align. & 0.838 & \textbf{0.936} & \textcolor{teal}{$\uparrow$ 11.7\%} \\
Set A: Instr. Faith.   & 0.841 & \textbf{0.946} & \textcolor{teal}{$\uparrow$ 12.5\%} \\
Set A: Spatial Logic   & 0.833 & \textbf{0.926} & \textcolor{teal}{$\uparrow$ 11.2\%} \\
Set A: \textbf{Composite} & 0.841 & \textbf{0.929} & \textcolor{teal}{$\uparrow$ 10.5\%} \\
\cmidrule(lr){1-4}
Set B: Instr. Faith.$^\dagger$ & 0.549 & \textbf{0.619} & \textcolor{teal}{$\uparrow$ 12.8\%} \\
Set B: \textbf{Composite}     & 0.662 & \textbf{0.703} & \textcolor{teal}{$\uparrow$ 6.2\%} \\
\midrule
\multicolumn{4}{l}{\textit{\textbf{Part II: Understanding Task (Accuracy \%)}}} \\
\midrule
Task I: Direct (Overall) & 14.0 & \textbf{20.0} & \textcolor{teal}{$\uparrow$ 42.9\%} \\
Task II: Select (Overall)& 9.5  & \textbf{10.5} & \textcolor{teal}{$\uparrow$ 10.5\%} \\
\bottomrule
\end{tabular}%
}
\label{tab:table3}
\end{table}

\section{Conclusion}
We introduced SVE-ASCII, a pioneering framework that unlocks the latent potential of LLMs for native visual expression through the isomorphic medium of ASCII art. By constructing the high-quality ASCIIArt-7K dataset via our ``Seed-and-Evolve" pipeline and establishing the comprehensive ASCIIArt-Bench, we have provided the community with robust resources for benchmarking symbolic visual processing. Our unified instruction-tuning strategy not only achieves state-of-the-art performance in generating high-fidelity symbolic art but also empirically validates the ``Cycle of Mutual Reinforcement". Crucially, we demonstrate that the generative rigor required to synthesize coherent spatial structures significantly enhances the model's visual comprehension. This work effectively bridges the syntax-semantics gap inherent in code-based approaches, proving that LLMs possess an intrinsic capacity for visual intelligence without reliance on external renderers. 

Building on this structural interpretability, we further envision SVE as a ``Visual Chain-of-Thought" to guide pixel-based diffusion models, leveraging the logical coherence of LLMs to align complex visual synthesis with textual prompts more effectively.


\bibliographystyle{icml2026}
\bibliography{example_paper}

\begin{thebibliography}{37}
\providecommand{\natexlab}[1]{#1}
\providecommand{\url}[1]{\texttt{#1}}
\expandafter\ifx\csname urlstyle\endcsname\relax
  \providecommand{\doi}[1]{doi: #1}\else
  \providecommand{\doi}{doi: \begingroup \urlstyle{rm}\Url}\fi

\bibitem[Bai et~al.(2025)Bai, Bao, Chen, Chen, Chen, Chen, Chen, Chen, Chen, Chen, Cui, Ding, Dong, Du, Du, Du, Du, Fan, Feng, Fu, Gao, Gao, Gao, Gao, Gu, Guan, Guo, Guo, Hu, Hao, He, He, He, Hong, Hu, Hu, Huang, Huang, Huang, Jiang, Jiang, Jin, Kang, Lai, Li, Li, Li, Li, Li, Li, Li, Li, Li, Lin, Lin, Lin, Liu, Liu, Liu, Liu, Liu, Liu, Liu, Liu, Liu, Liu, Liu, Liu, Liu, Liu, Liu, Lu, Lu, Ma, Ma, Ma, Mao, Mei, Men, Miao, Pan, Peng, Qin, Qu, Shang, Shi, Shi, Song, Su, Su, Sun, Sung, Tang, Tao, Teng, Wang, Wang, Wang, and Wang]{DBLP:journals/corr/abs-2507-20534}
Bai, Y., Bao, Y., Chen, G., Chen, J., Chen, N., Chen, R., Chen, Y., Chen, Y., Chen, Y., Chen, Z., Cui, J., Ding, H., Dong, M., Du, A., Du, C., Du, D., Du, Y., Fan, Y., Feng, Y., Fu, K., Gao, B., Gao, H., Gao, P., Gao, T., Gu, X., Guan, L., Guo, H., Guo, J., Hu, H., Hao, X., He, T., He, W., He, W., Hong, C., Hu, Y., Hu, Z., Huang, W., Huang, Z., Huang, Z., Jiang, T., Jiang, Z., Jin, X., Kang, Y., Lai, G., Li, C., Li, F., Li, H., Li, M., Li, W., Li, Y., Li, Y., Li, Z., Li, Z., Lin, H., Lin, X., Lin, Z., Liu, C., Liu, C., Liu, H., Liu, J., Liu, J., Liu, L., Liu, S., Liu, T.~Y., Liu, T., Liu, W., Liu, Y., Liu, Y., Liu, Y., Liu, Y., Liu, Z., Lu, E., Lu, L., Ma, S., Ma, X., Ma, Y., Mao, S., Mei, J., Men, X., Miao, Y., Pan, S., Peng, Y., Qin, R., Qu, B., Shang, Z., Shi, L., Shi, S., Song, F., Su, J., Su, Z., Sun, X., Sung, F., Tang, H., Tao, J., Teng, Q., Wang, C., Wang, D., Wang, F., and Wang, H.
\newblock Kimi {K2:} open agentic intelligence.
\newblock \emph{CoRR}, abs/2507.20534, 2025.
\newblock \doi{10.48550/ARXIV.2507.20534}.
\newblock URL \url{https://doi.org/10.48550/arXiv.2507.20534}.

\bibitem[Carlier et~al.(2020)Carlier, Danelljan, Alahi, and Timofte]{DeepSVG}
Carlier, A., Danelljan, M., Alahi, A., and Timofte, R.
\newblock Deepsvg: {A} hierarchical generative network for vector graphics animation.
\newblock In Larochelle, H., Ranzato, M., Hadsell, R., Balcan, M., and Lin, H. (eds.), \emph{Advances in Neural Information Processing Systems 33: Annual Conference on Neural Information Processing Systems 2020, NeurIPS 2020, December 6-12, 2020, virtual}, 2020.
\newblock URL \url{https://proceedings.neurips.cc/paper/2020/hash/bcf9d6bd14a2095866ce8c950b702341-Abstract.html}.

\bibitem[Chakrabarty et~al.(2023)Chakrabarty, Saakyan, Winn, Panagopoulou, Yang, Apidianaki, and Muresan]{DBLP:conf/acl/ChakrabartySWPY23}
Chakrabarty, T., Saakyan, A., Winn, O., Panagopoulou, A., Yang, Y., Apidianaki, M., and Muresan, S.
\newblock I spy a metaphor: Large language models and diffusion models co-create visual metaphors.
\newblock In Rogers, A., Boyd{-}Graber, J.~L., and Okazaki, N. (eds.), \emph{Findings of the Association for Computational Linguistics: {ACL} 2023, Toronto, Canada, July 9-14, 2023}, volume {ACL} 2023 of \emph{Findings of {ACL}}, pp.\  7370--7388. Association for Computational Linguistics, 2023.
\newblock \doi{10.18653/V1/2023.FINDINGS-ACL.465}.
\newblock URL \url{https://doi.org/10.18653/v1/2023.findings-acl.465}.

\bibitem[Chen et~al.(2023)Chen, Wang, Changpinyo, Piergiovanni, Padlewski, Salz, Goodman, Grycner, Mustafa, Beyer, Kolesnikov, Puigcerver, Ding, Rong, Akbari, Mishra, Xue, Thapliyal, Bradbury, and Kuo]{DBLP:conf/iclr/Chen0CPPSGGMB0P23}
Chen, X., Wang, X., Changpinyo, S., Piergiovanni, A.~J., Padlewski, P., Salz, D., Goodman, S., Grycner, A., Mustafa, B., Beyer, L., Kolesnikov, A., Puigcerver, J., Ding, N., Rong, K., Akbari, H., Mishra, G., Xue, L., Thapliyal, A.~V., Bradbury, J., and Kuo, W.
\newblock Pali: {A} jointly-scaled multilingual language-image model.
\newblock In \emph{The Eleventh International Conference on Learning Representations, {ICLR} 2023, Kigali, Rwanda, May 1-5, 2023}. OpenReview.net, 2023.
\newblock URL \url{https://openreview.net/forum?id=mWVoBz4W0u}.

\bibitem[Coumar \& Kingston(2025)Coumar and Kingston]{DBLP:journals/corr/abs-2503-14375}
Coumar, S. and Kingston, Z.
\newblock Evaluating machine learning approaches for {ASCII} art generation.
\newblock \emph{CoRR}, abs/2503.14375, 2025.
\newblock \doi{10.48550/ARXIV.2503.14375}.
\newblock URL \url{https://doi.org/10.48550/arXiv.2503.14375}.

\bibitem[de~Jong et~al.(2006)de~Jong, van Hijum, Bijlsma, Kok, and Kuipers]{de2006bagel}
de~Jong, A., van Hijum, S.~A., Bijlsma, J.~J., Kok, J., and Kuipers, O.~P.
\newblock Bagel: a web-based bacteriocin genome mining tool.
\newblock \emph{Nucleic acids research}, 34\penalty0 (suppl\_2):\penalty0 W273--W279, 2006.

\bibitem[DeepSeek{-}AI(2025)]{DBLP:journals/corr/abs-2512-02556}
DeepSeek{-}AI.
\newblock Deepseek-v3.2: Pushing the frontier of open large language models.
\newblock \emph{CoRR}, abs/2512.02556, 2025.
\newblock \doi{10.48550/ARXIV.2512.02556}.
\newblock URL \url{https://doi.org/10.48550/arXiv.2512.02556}.

\bibitem[Deng \& Li(2025)Deng and Li]{DBLP:journals/corr/abs-2509-21404}
Deng, X. and Li, H.
\newblock How large language models need symbolism.
\newblock \emph{CoRR}, abs/2509.21404, 2025.
\newblock \doi{10.48550/ARXIV.2509.21404}.
\newblock URL \url{https://doi.org/10.48550/arXiv.2509.21404}.

\bibitem[Esser et~al.(2021)Esser, Rombach, and Ommer]{DBLP:conf/cvpr/EsserRO21}
Esser, P., Rombach, R., and Ommer, B.
\newblock Taming transformers for high-resolution image synthesis.
\newblock In \emph{{IEEE} Conference on Computer Vision and Pattern Recognition, {CVPR} 2021, virtual, June 19-25, 2021}, pp.\  12873--12883. Computer Vision Foundation / {IEEE}, 2021.
\newblock \doi{10.1109/CVPR46437.2021.01268}.
\newblock URL \url{https://openaccess.thecvf.com/content/CVPR2021/html/Esser\_Taming\_Transformers\_for\_High-Resolution\_Image\_Synthesis\_CVPR\_2021\_paper.html}.

\bibitem[Feng et~al.(2020)Feng, Guo, Tang, Duan, Feng, Gong, Shou, Qin, Liu, Jiang, and Zhou]{DBLP:conf/emnlp/FengGTDFGS0LJZ20}
Feng, Z., Guo, D., Tang, D., Duan, N., Feng, X., Gong, M., Shou, L., Qin, B., Liu, T., Jiang, D., and Zhou, M.
\newblock Codebert: {A} pre-trained model for programming and natural languages.
\newblock In Cohn, T., He, Y., and Liu, Y. (eds.), \emph{Findings of the Association for Computational Linguistics: {EMNLP} 2020, Online Event, 16-20 November 2020}, volume {EMNLP} 2020 of \emph{Findings of {ACL}}, pp.\  1536--1547. Association for Computational Linguistics, 2020.
\newblock \doi{10.18653/V1/2020.FINDINGS-EMNLP.139}.
\newblock URL \url{https://doi.org/10.18653/v1/2020.findings-emnlp.139}.

\bibitem[Jia et~al.(2024)Jia, Yue, Huang, Qin, Liu, Lin, You, and Zhai]{ASCIIEval2024}
Jia, Q., Yue, X., Huang, S., Qin, Z., Liu, Y., Lin, B.~Y., You, Y., and Zhai, G.
\newblock Asciieval: Benchmarking models' visual perception in text strings via ascii art, 2024.
\newblock URL \url{https://arxiv.org/abs/2410.01733}.
\newblock arXiv:2410.01733.

\bibitem[Koh et~al.(2023)Koh, Fried, and Salakhutdinov]{DBLP:conf/nips/KohFS23}
Koh, J.~Y., Fried, D., and Salakhutdinov, R.
\newblock Generating images with multimodal language models.
\newblock In Oh, A., Naumann, T., Globerson, A., Saenko, K., Hardt, M., and Levine, S. (eds.), \emph{Advances in Neural Information Processing Systems 36: Annual Conference on Neural Information Processing Systems 2023, NeurIPS 2023, New Orleans, LA, USA, December 10 - 16, 2023}, 2023.
\newblock URL \url{http://papers.nips.cc/paper\_files/paper/2023/hash/43a69d143273bd8215578bde887bb552-Abstract-Conference.html}.

\bibitem[Li et~al.(2023)Li, Prabhudesai, Duggal, Brown, and Pathak]{DBLP:conf/iccv/LiPDBP23}
Li, A.~C., Prabhudesai, M., Duggal, S., Brown, E., and Pathak, D.
\newblock Your diffusion model is secretly a zero-shot classifier.
\newblock In \emph{{IEEE/CVF} International Conference on Computer Vision, {ICCV} 2023, Paris, France, October 1-6, 2023}, pp.\  2206--2217. {IEEE}, 2023.
\newblock \doi{10.1109/ICCV51070.2023.00210}.
\newblock URL \url{https://doi.org/10.1109/ICCV51070.2023.00210}.

\bibitem[Lu et~al.(2024)Lu, Clark, Lee, Zhang, Khosla, Marten, Hoiem, and Kembhavi]{DBLP:conf/cvpr/LuCL0KMHK24}
Lu, J., Clark, C., Lee, S., Zhang, Z., Khosla, S., Marten, R., Hoiem, D., and Kembhavi, A.
\newblock Unified-io 2: Scaling autoregressive multimodal models with vision, language, audio, and action.
\newblock In \emph{{IEEE/CVF} Conference on Computer Vision and Pattern Recognition, {CVPR} 2024, Seattle, WA, USA, June 16-22, 2024}, pp.\  26429--26445. {IEEE}, 2024.
\newblock \doi{10.1109/CVPR52733.2024.02497}.
\newblock URL \url{https://doi.org/10.1109/CVPR52733.2024.02497}.

\bibitem[Luo et~al.(2025)Luo, Fu, Peguero, Malik, Patil, Lin, Overborg, Sarmiento, and Zhu]{ASCIIBench}
Luo, K., Fu, M., Peguero, J., Malik, H., Patil, A., Lin, J., Overborg, M.~V., Sarmiento, R., and Zhu, K.
\newblock Asciibench: Evaluating language-model-based understanding of visually-oriented text.
\newblock \emph{CoRR}, abs/2512.04125, 2025.
\newblock \doi{10.48550/ARXIV.2512.04125}.
\newblock URL \url{https://doi.org/10.48550/arXiv.2512.04125}.

\bibitem[Podell et~al.(2024)Podell, English, Lacey, Blattmann, Dockhorn, M{\"{u}}ller, Penna, and Rombach]{SDXL}
Podell, D., English, Z., Lacey, K., Blattmann, A., Dockhorn, T., M{\"{u}}ller, J., Penna, J., and Rombach, R.
\newblock {SDXL:} improving latent diffusion models for high-resolution image synthesis.
\newblock In \emph{The Twelfth International Conference on Learning Representations, {ICLR} 2024, Vienna, Austria, May 7-11, 2024}. OpenReview.net, 2024.
\newblock URL \url{https://openreview.net/forum?id=di52zR8xgf}.

\bibitem[Priyanshu et~al.(2024)Priyanshu, Maurya, and Hong]{Claude}
Priyanshu, A., Maurya, Y., and Hong, Z.
\newblock {AI} governance and accountability: An analysis of anthropic's claude.
\newblock \emph{CoRR}, abs/2407.01557, 2024.
\newblock \doi{10.48550/ARXIV.2407.01557}.
\newblock URL \url{https://doi.org/10.48550/arXiv.2407.01557}.

\bibitem[Rodr{\'{\i}}guez et~al.(2025)Rodr{\'{\i}}guez, Puri, Agarwal, Laradji, Rodr{\'{\i}}guez, Rajeswar, V{\'{a}}zquez, Pal, and Pedersoli]{StarVector}
Rodr{\'{\i}}guez, J.~A., Puri, A., Agarwal, S., Laradji, I.~H., Rodr{\'{\i}}guez, P., Rajeswar, S., V{\'{a}}zquez, D., Pal, C., and Pedersoli, M.
\newblock Starvector: Generating scalable vector graphics code from images and text.
\newblock In \emph{{IEEE/CVF} Conference on Computer Vision and Pattern Recognition, {CVPR} 2025, Nashville, TN, USA, June 11-15, 2025}, pp.\  16175--16186. Computer Vision Foundation / {IEEE}, 2025.
\newblock \doi{10.1109/CVPR52734.2025.01508}.
\newblock URL \url{https://openaccess.thecvf.com/content/CVPR2025/html/Rodriguez\_StarVector\_Generating\_Scalable\_Vector\_Graphics\_Code\_from\_Images\_and\_Text\_CVPR\_2025\_paper.html}.

\bibitem[Saharia et~al.(2022)Saharia, Chan, Saxena, Li, Whang, Denton, Ghasemipour, Lopes, Ayan, Salimans, Ho, Fleet, and Norouzi]{Imagen}
Saharia, C., Chan, W., Saxena, S., Li, L., Whang, J., Denton, E.~L., Ghasemipour, S. K.~S., Lopes, R.~G., Ayan, B.~K., Salimans, T., Ho, J., Fleet, D.~J., and Norouzi, M.
\newblock Photorealistic text-to-image diffusion models with deep language understanding.
\newblock In Koyejo, S., Mohamed, S., Agarwal, A., Belgrave, D., Cho, K., and Oh, A. (eds.), \emph{Advances in Neural Information Processing Systems 35: Annual Conference on Neural Information Processing Systems 2022, NeurIPS 2022, New Orleans, LA, USA, November 28 - December 9, 2022}, 2022.
\newblock URL \url{http://papers.nips.cc/paper\_files/paper/2022/hash/ec795aeadae0b7d230fa35cbaf04c041-Abstract-Conference.html}.

\bibitem[Seghair et~al.(2024)Seghair, Besbes, Abdellatif, and Bihiri]{DBLP:conf/bigdataconf/SeghairBAB24}
Seghair, T., Besbes, O., Abdellatif, T., and Bihiri, S.
\newblock {VQ-VGAE:} vector quantized variational graph auto-encoder for unsupervised anomaly detection.
\newblock In Ding, W., Lu, C., Wang, F., Di, L., Wu, K., Huan, J., Nambiar, R., Li, J., Ilievski, F., Baeza{-}Yates, R., and Hu, X. (eds.), \emph{{IEEE} International Conference on Big Data, BigData 2024, Washington, DC, USA, December 15-18, 2024}, pp.\  2370--2375. {IEEE}, 2024.
\newblock \doi{10.1109/BIGDATA62323.2024.10825598}.
\newblock URL \url{https://doi.org/10.1109/BigData62323.2024.10825598}.

\bibitem[Song(2022)]{song2022cliptexture}
Song, Y.
\newblock Cliptexture: Text-driven texture synthesis.
\newblock In \emph{Proceedings of the 30th ACM International Conference on Multimedia}, pp.\  5468--5476, 2022.

\bibitem[Song \& Zhang(2022)Song and Zhang]{song2022clipfont}
Song, Y. and Zhang, Y.
\newblock Clipfont: Text guided vector wordart generation.
\newblock In \emph{BMVC}, pp.\  543, 2022.

\bibitem[Song et~al.(2023)Song, Shao, Chen, Zhang, Jing, and Li]{song2023clipvg}
Song, Y., Shao, X., Chen, K., Zhang, W., Jing, Z., and Li, M.
\newblock Clipvg: Text-guided image manipulation using differentiable vector graphics.
\newblock In \emph{Proceedings of the AAAI conference on artificial intelligence}, volume~37, pp.\  2312--2320, 2023.

\bibitem[Song et~al.(2025)Song, Chen, and Shou]{song2025layertracer}
Song, Y., Chen, D., and Shou, M.~Z.
\newblock Layertracer: Cognitive-aligned layered svg synthesis via diffusion transformer.
\newblock \emph{arXiv preprint arXiv:2502.01105}, 2025.

\bibitem[Sur{\'{\i}}s et~al.(2023)Sur{\'{\i}}s, Menon, and Vondrick]{ViperGPT}
Sur{\'{\i}}s, D., Menon, S., and Vondrick, C.
\newblock Vipergpt: Visual inference via python execution for reasoning.
\newblock In \emph{{IEEE/CVF} International Conference on Computer Vision, {ICCV} 2023, Paris, France, October 1-6, 2023}, pp.\  11854--11864. {IEEE}, 2023.
\newblock \doi{10.1109/ICCV51070.2023.01092}.
\newblock URL \url{https://doi.org/10.1109/ICCV51070.2023.01092}.

\bibitem[Team(2023)]{DBLP:journals/corr/abs-2312-11805}
Team, G.
\newblock Gemini: {A} family of highly capable multimodal models.
\newblock \emph{CoRR}, abs/2312.11805, 2023.
\newblock \doi{10.48550/ARXIV.2312.11805}.
\newblock URL \url{https://doi.org/10.48550/arXiv.2312.11805}.

\bibitem[Tian et~al.(2023)Tian, Fan, Isola, Chang, and Krishnan]{DBLP:conf/nips/TianFICK23}
Tian, Y., Fan, L., Isola, P., Chang, H., and Krishnan, D.
\newblock Stablerep: Synthetic images from text-to-image models make strong visual representation learners.
\newblock In Oh, A., Naumann, T., Globerson, A., Saenko, K., Hardt, M., and Levine, S. (eds.), \emph{Advances in Neural Information Processing Systems 36: Annual Conference on Neural Information Processing Systems 2023, NeurIPS 2023, New Orleans, LA, USA, December 10 - 16, 2023}, 2023.
\newblock URL \url{http://papers.nips.cc/paper\_files/paper/2023/hash/971f1e59cd956cc094da4e2f78c6ea7c-Abstract-Conference.html}.

\bibitem[Tong et~al.(2025)Tong, Gu, Lou, Fan, Zou, Wu, Ye, and Li]{DBLP:journals/corr/abs-2512-16584}
Tong, J., Gu, J., Lou, Y., Fan, L., Zou, Y., Wu, Y., Ye, J., and Li, R.
\newblock Sketch-in-latents: Eliciting unified reasoning in mllms.
\newblock \emph{CoRR}, abs/2512.16584, 2025.
\newblock \doi{10.48550/ARXIV.2512.16584}.
\newblock URL \url{https://doi.org/10.48550/arXiv.2512.16584}.

\bibitem[Wang et~al.(2021)Wang, Wang, Joty, and Hoi]{DBLP:conf/emnlp/0034WJH21}
Wang, Y., Wang, W., Joty, S.~R., and Hoi, S. C.~H.
\newblock Codet5: Identifier-aware unified pre-trained encoder-decoder models for code understanding and generation.
\newblock In Moens, M., Huang, X., Specia, L., and Yih, S.~W. (eds.), \emph{Proceedings of the 2021 Conference on Empirical Methods in Natural Language Processing, {EMNLP} 2021, Virtual Event / Punta Cana, Dominican Republic, 7-11 November, 2021}, pp.\  8696--8708. Association for Computational Linguistics, 2021.
\newblock \doi{10.18653/V1/2021.EMNLP-MAIN.685}.
\newblock URL \url{https://doi.org/10.18653/v1/2021.emnlp-main.685}.

\bibitem[Wu et~al.(2023{\natexlab{a}})Wu, Yin, Qi, Wang, Tang, and Duan]{VisualChatGPT}
Wu, C., Yin, S., Qi, W., Wang, X., Tang, Z., and Duan, N.
\newblock Visual chatgpt: Talking, drawing and editing with visual foundation models.
\newblock \emph{CoRR}, abs/2303.04671, 2023{\natexlab{a}}.
\newblock \doi{10.48550/ARXIV.2303.04671}.
\newblock URL \url{https://doi.org/10.48550/arXiv.2303.04671}.

\bibitem[Wu et~al.(2023{\natexlab{b}})Wu, Su, Ma, and Liao]{IconShop}
Wu, R., Su, W., Ma, K., and Liao, J.
\newblock Iconshop: Text-guided vector icon synthesis with autoregressive transformers.
\newblock \emph{{ACM} Trans. Graph.}, 42\penalty0 (6):\penalty0 230:1--230:14, 2023{\natexlab{b}}.
\newblock \doi{10.1145/3618364}.
\newblock URL \url{https://doi.org/10.1145/3618364}.

\bibitem[Wu et~al.(2025)Wu, Su, and Liao]{Chat2SVG}
Wu, R., Su, W., and Liao, J.
\newblock Chat2svg: Vector graphics generation with large language models and image diffusion models.
\newblock In \emph{{IEEE/CVF} Conference on Computer Vision and Pattern Recognition, {CVPR} 2025, Nashville, TN, USA, June 11-15, 2025}, pp.\  23690--23700. Computer Vision Foundation / {IEEE}, 2025.
\newblock \doi{10.1109/CVPR52734.2025.02206}.
\newblock URL \url{https://openaccess.thecvf.com/content/CVPR2025/html/Wu\_Chat2SVG\_Vector\_Graphics\_Generation\_with\_Large\_Language\_Models\_and\_Image\_CVPR\_2025\_paper.html}.

\bibitem[Xie et~al.(2025)Xie, Mao, Bai, Zhang, Wang, Lin, Gu, Chen, Yang, and Shou]{Show-o}
Xie, J., Mao, W., Bai, Z., Zhang, D.~J., Wang, W., Lin, K.~Q., Gu, Y., Chen, Z., Yang, Z., and Shou, M.~Z.
\newblock Show-o: One single transformer to unify multimodal understanding and generation.
\newblock In \emph{The Thirteenth International Conference on Learning Representations, {ICLR} 2025, Singapore, April 24-28, 2025}. OpenReview.net, 2025.
\newblock URL \url{https://openreview.net/forum?id=o6Ynz6OIQ6}.

\bibitem[Yang et~al.(2024)Yang, Yang, Zhang, Hui, Zheng, Yu, Li, Liu, Huang, Wei, Lin, Yang, Tu, Zhang, Yang, Yang, Zhou, Lin, Dang, Lu, Bao, Yang, Yu, Li, Xue, Zhang, Zhu, Men, Lin, Li, Xia, Ren, Ren, Fan, Su, Zhang, Wan, Liu, Cui, Zhang, and Qiu]{Qwen2.5}
Yang, A., Yang, B., Zhang, B., Hui, B., Zheng, B., Yu, B., Li, C., Liu, D., Huang, F., Wei, H., Lin, H., Yang, J., Tu, J., Zhang, J., Yang, J., Yang, J., Zhou, J., Lin, J., Dang, K., Lu, K., Bao, K., Yang, K., Yu, L., Li, M., Xue, M., Zhang, P., Zhu, Q., Men, R., Lin, R., Li, T., Xia, T., Ren, X., Ren, X., Fan, Y., Su, Y., Zhang, Y., Wan, Y., Liu, Y., Cui, Z., Zhang, Z., and Qiu, Z.
\newblock Qwen2.5 technical report.
\newblock \emph{CoRR}, abs/2412.15115, 2024.
\newblock \doi{10.48550/ARXIV.2412.15115}.
\newblock URL \url{https://doi.org/10.48550/arXiv.2412.15115}.

\bibitem[Yang et~al.(2025)Yang, Cheng, Chen, Zeng, Zhang, Wang, Yu, Ma, and Jiang]{OmniSVG}
Yang, Y., Cheng, W., Chen, S., Zeng, X., Zhang, J., Wang, L., Yu, G., Ma, X., and Jiang, Y.
\newblock Omnisvg: {A} unified scalable vector graphics generation model.
\newblock \emph{CoRR}, abs/2504.06263, 2025.
\newblock \doi{10.48550/ARXIV.2504.06263}.
\newblock URL \url{https://doi.org/10.48550/arXiv.2504.06263}.

\bibitem[Ye et~al.(2025)Ye, Liu, and Song]{ye2025loom}
Ye, M., Liu, J., and Song, Y.
\newblock Loom: Diffusion-transformer for interleaved generation.
\newblock \emph{arXiv preprint arXiv:2512.18254}, 2025.

\bibitem[Zhang et~al.(2026)Zhang, Bai, Wang, and Song]{zhang2026sigma}
Zhang, X., Bai, Z., Wang, H., and Song, Y.
\newblock Sigma: Selective-interleaved generation with multi-attribute tokens.
\newblock \emph{arXiv preprint arXiv:2602.07564}, 2026.

\end{thebibliography}


\appendix
\onecolumn
\section{More Results of Ours and Baselines}
\begin{figure}[htbp]
    \centering
    \includegraphics[width=\textwidth]{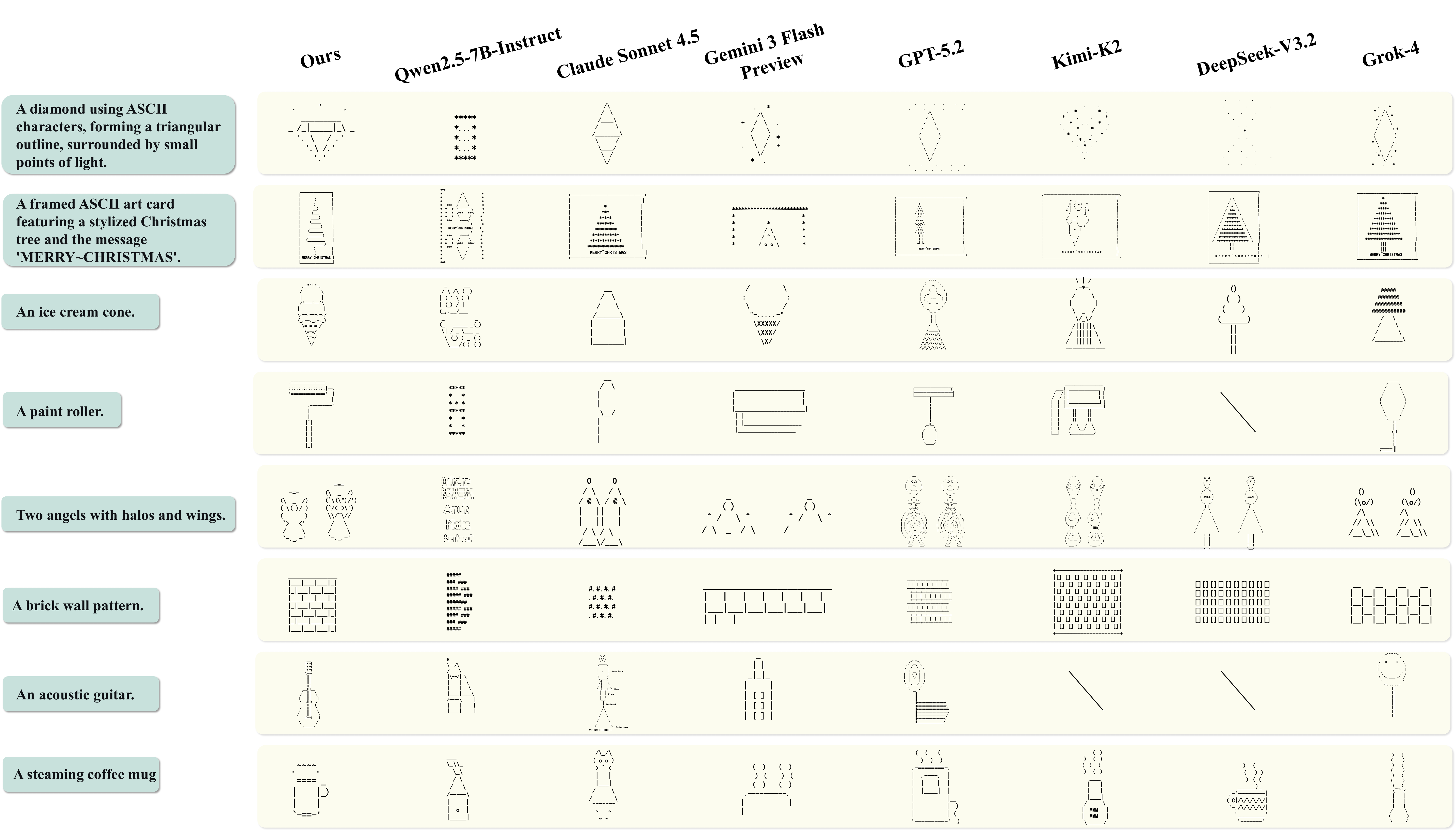}
    \caption{More results of ours and baselines.}
    \label{fig:placeholder}
\end{figure}

\section{Task Prompt Templates and Configurations of Benchmark}
\label{benchmark-task}
\subsection{ASCII Art Generation (Text-to-Art)}

For the generation task, the system enforces a strict output format to ensure parser compatibility.

\begin{promptbox}{System Prompt (Generation)}
You are an ASCII art generation expert. \\
Your goal is to create precise and structurally correct ASCII art based on user requests. \\
Output format: \\
\texttt{<art>} \\
\text{[Final ASCII Art]} \\
\texttt{</art>}
\end{promptbox}

\begin{promptbox}{User Prompt (Generation)}
\{instruction\}
\end{promptbox}

\subsection{ASCII Art Understanding (Art-to-Text)}
\noindent\begin{minipage}{\linewidth}
    We evaluate understanding through both Direct Understanding and Category Selection.
    
    \vspace{0.5em} 

    \begin{promptbox}{Understanding Task Configuration}
    \small 
    
    \textbf{System Prompt:} \\
    You are an expert in interpreting ASCII art. Describe the content of the following ASCII illustration precisely.
    
    \vspace{0.8em} \hrule \vspace{0.8em} 
    
    \textbf{User Prompt (Direct Understanding):} \\
    What is depicted in the following ASCII art? \\
    \texttt{<art>} \\
    \{ascii\_art\} \\
    \texttt{</art>}
    
    \vspace{0.8em} \hrule \vspace{0.8em} 
    
    \textbf{User Prompt (Category Selection):} \\
    What is depicted in the following ASCII art? \\
    \texttt{<art>} \\
    \{ascii\_art\} \\
    \texttt{</art>} \\
    Please select the correct category from the options below. \\
    Options: \\
    \{options\_list\} \\
    Please respond with only the category name (e.g., "Cat", "Wolf").
    \end{promptbox}

\end{minipage}

\subsection{Data Configurations}

\noindent\begin{minipage}{\linewidth}
    \centering
    \vspace{0.5em} 
    
    \captionof{table}{\textbf{Statistics of the ASCIIArt-Bench.} We report the number of samples, unique categories, and structural statistics (average dimensions and area) for each subset. The Understanding task evaluates both Direct Understanding and Category Selection on a shared dataset using distinct prompt formulations.}
    \label{tab:benchmark_stats}
    
    \setlength{\tabcolsep}{5pt} 
    
    \begin{tabular}{l|c|c|c|c}
        \toprule
        \textbf{Task / Subset} & \textbf{Samples} & \textbf{Categories} & \textbf{Avg. Size} ($W \times H$) & \textbf{Area} (Med. / Max.) \\
        \midrule
        
        \multicolumn{5}{l}{\textbf{Generation Task}} \\
        \cmidrule(lr){1-5} 
        Recall & 100 & 72 & \multirow{2}{*}{$18.0 \times 10.0$} & \multirow{2}{*}{151.5 / 784} \\
        \cmidrule(lr){1-3} 
        Generalization & 100 & 67 & & \\
        \midrule
        
        \multicolumn{5}{l}{\textbf{Understanding Task} (Direct Understanding and Category Selection)} \\
        \cmidrule(lr){1-5}
        Seen & 100 & 60 & \multirow{2}{*}{$27.8 \times 8.2$} & \multirow{2}{*}{174.5 / 1100} \\
        \cmidrule(lr){1-3}
        Unseen & 100 & 82 & & \\
        \bottomrule
    \end{tabular}
    
    \vspace{1em} 
\end{minipage}

\section{LLM-as-Judge Evaluation Templates}
\label{LLM-as-Judge}

\noindent\begin{minipage}{\linewidth}

    \subsection{Multi-Dimensional Scoring for Generation}
    The judge evaluates generated art based on a composite score of five dimensions.
    
    \vspace{0.3em} 

    \begin{promptbox}{Gemini Judge Prompt (Generation)}
    \small 
    
    You are a strict expert ASCII Art Critic. Your job is to verify if the generated art MATCHES the user's request: "\{instruction\}".
    
    \textbf{Inputs Provided:}
    \begin{itemize}[leftmargin=1.5em, itemsep=0pt, topsep=2pt, parsep=0pt]
        \item 1. The rendered IMAGE (Visual check).
        \item 2. The raw TEXT characters (Structure check).
    \end{itemize}
    
    \textbf{Evaluate on these 5 dimensions (Score 0.0 to 1.0):}
    \begin{itemize}[leftmargin=1.5em, itemsep=5pt, topsep=2pt, parsep=0pt]
        \item \textbf{SA (Semantic Alignment)}: Visually looks like the main object requested?
        \item \textbf{IF (Instruction Faithfulness)}: Follows constraints (e.g., "facing left")? 
        \item \textbf{SC (Structural Coherence)}: Lines connected? Shape closed and logical?
        \item \textbf{SL (Spatial Logic)}: Parts (head, legs) in correct relative positions?
        \item \textbf{CE (Character Efficiency)}: Clean? Penalize spamming/grid-filling.
    \end{itemize}
    
    \textbf{CRITICAL INSTRUCTIONS:}
    \begin{enumerate}[leftmargin=1.5em, itemsep=2pt, topsep=2pt, parsep=0pt]
        \item \textbf{TRANSCRIPTION FIRST}: MUST "read" the art from the image first.
        \begin{itemize}[leftmargin=1em, label=-, itemsep=0pt]
            \item If text: type exact letters; If object: describe orientation objectively.
        \end{itemize}
        \item \textbf{COMPARE}: Transcription vs. Request (e.g., User "left" vs. You see "right" $\to$ low IF).
    \end{enumerate}
    
    Output JSON format ONLY: \\
    \texttt{\{ \\
    \hspace*{1.5em}"SA": <float>, "IF": <float>, "SC": <float>, \\
    \hspace*{1.5em}"SL": <float>, "CE": <float>, \\
    \hspace*{1.5em}"reasoning": "<short explanation>" \\
    \}}
    \end{promptbox}

\end{minipage} 
\vspace{0.5em} 

\noindent\begin{minipage}{\linewidth}

    \subsection{Semantic Equivalence for Understanding}
    For understanding tasks, the judge determines if the model's description matches the ground truth intent.
    
    \vspace{0.5em} 

    \begin{promptbox}{Gemini Judge Prompt (Understanding)}
    \small 
    
    You are an intelligent evaluator.
    \textbf{Task}: Determine if the Model's Output conveys the \textbf{same meaning} as the Ground Truth.
    
    \vspace{2pt}
    \textbf{Inputs:}
    \begin{itemize}[leftmargin=1.5em, itemsep=3pt, topsep=2pt, parsep=0pt]
        \item \textbf{Ground Truth}: \{ground\_truth\}
        \item \textbf{Model's Output}: \{model\_output\}
    \end{itemize}
    
    \textbf{Evaluation Criteria:}
    \begin{enumerate}[leftmargin=1.5em, itemsep=2pt, topsep=2pt, parsep=0pt]
        \item \textbf{Semantic Equivalence}: Focus on meaning/intent, not exact wording.
        \item \textbf{Allow Synonyms}: Be flexible with paraphrasing (e.g., "taxi" matches "cab").
        \item \textbf{Ignore Noise}: Disregard punctuation, casing, typos, or polite fillers.
    \end{enumerate}
    
    Output JSON format ONLY: \\
    \texttt{\{ "is\_correct": <bool>, "confidence": <float>, \\
    \hspace*{1em}"reasoning": "<Very brief explanation>" \}}
    \end{promptbox}

\end{minipage}
\vspace{1em} 


\end{document}